\documentclass{article}

\usepackage{PRIMEarxiv}
\usepackage[utf8]{inputenc} 
\usepackage[T1]{fontenc}    
\usepackage{hyperref}       
\usepackage{url}            
\usepackage{booktabs}       
\usepackage{amsfonts}       
\usepackage{nicefrac}       
\usepackage{microtype}      
\usepackage{lipsum}
\usepackage{fancyhdr}       
\usepackage{graphicx}       
\graphicspath{{media/}}     
\usepackage{amsmath}

\title{Brain-to-Text Benchmark '24: Lessons Learned}

\author{
  Francis R. Willett \\
  Department of Neurosurgery \\
  Stanford University \\
  Competition Organizer\\
  \texttt{fwillett@stanford.edu} \\
  \And
  Jingyuan Li$^*$, Trung Le$^*$, Chaofei Fan$^\ddagger$, Mingfei Chen$^*$, Eli Shlizerman$^{*+}$ \\
  \text{*} Department of Electrical \& Computer Engineering,
  University of Washington \\
 $\ddagger$ Department of Computer Science,
  Stanford University\\
  + Department of Applied Mathematics,
  University of Washington \\
  1st Place Entry\\
  \And
  Yue Chen, Xin Zheng, Tatsuo S. Okubo \\
  Beijing Institute for Brain Research, Chinese Academy of Medical Sciences \\ Peking Union Medical College\\
  Chinese Institute for Brain Research, Beijing (CIBR)\\
  2nd Place Entry\\
  \And
  Tyler Benster \\
  Neurosciences Phd Program \\
  Stanford University \\
  3rd Place Entry\\
  \And
  Hyun Dong Lee, Maxwell Kounga, E. Kelly Buchanan, David Zoltowski, Scott W. Linderman \\
  Wu Tsai Neurosciences Institute and Department of Statistics \\
  Stanford University \\
  4th Place Entry \\
  \And
  Jaimie M. Henderson \\
  Department of Neurosurgery \\
  Stanford University \\
  Competition Organizer\\
  \texttt{henderj@stanford.edu} \\
}

\begin{document}
\maketitle
\begin{abstract}
Speech brain-computer interfaces aim to decipher what a person is trying to say from neural activity alone, restoring communication to people with paralysis who have lost the ability to speak intelligibly. \href{https://eval.ai/web/challenges/challenge-page/2099/overview}{The Brain-to-Text Benchmark '24} and associated competition was created to foster the advancement of decoding algorithms that convert neural activity to text. Here, we summarize the lessons learned from the competition ending on June 1, 2024 (the top 4 entrants also presented their experiences in a \href{https://www.youtube.com/watch?v=QzI2mr9OK_M&t=3s}{recorded webinar}). The largest improvements in accuracy were achieved using an ensembling approach, where the output of multiple independent decoders was merged using a fine-tuned large language model (an approach used by all 3 top entrants). Performance gains were also found by improving how the baseline recurrent neural network (RNN) model was trained, including by optimizing learning rate scheduling and by using a diphone training objective. Improving upon the model architecture itself proved more difficult, however, with attempts to use deep state space models or transformers not yet appearing to offer a benefit over the RNN baseline. The benchmark will remain open indefinitely to support further work towards increasing the accuracy of brain-to-text algorithms. 
\end{abstract}

\section{Introduction}
Recent work has shown impressive demonstrations of speech brain-computer interfaces (BCIs) that can restore communication to people with paralysis by deciphering the neural correlates of attempted speech and converting them to text \cite{card2024accurate,metzger2023high,willett2023high}. In the latest demonstration \cite{card2024accurate}, performance reached a level sufficient to restore natural conversation to the study participant, who now uses it daily to engage in his job and talk to his family, friends and caregivers. Despite this success, neural speech decoders still routinely make errors, impeding communication (particularly if the incorrect word is a key content word of the sentence). Increasing the accuracy of brain-to-text algorithms will make it more likely that speech BCIs can be clinically translated. 

To this end, we created the \href{https://eval.ai/web/challenges/challenge-page/2099/overview}{Brain-to-Text Benchmark '24} concurrent with the publication of our initial work and public release of the associated dataset \cite{willett2023high}. The Brain-to-Text Benchmark '24 measures the performance of neural decoding algorithms on a held-out, private dataset of 1200 sentences using the word error rate metric. We hosted a competition associated with the benchmark which officially ended on June 1, 2024. At that time, 4 entrants had improved upon our PyTorch baseline which implemented the algorithm described in our original work \cite{willett2023high}. Our baseline algorithm consisted of a recurrent neural network decoder trained to map neural activity to phoneme logits (using the connectionist temporal classification loss \cite{graves2006connectionist}), followed by a language model decoder that converted the phoneme logits into text. The language model decoding used two stages: a 5-gram hypothesis generating stage, followed by a large language model rescoring stage. The baseline algorithm achieved a 9.7\% word error rate, while the top entrant achieved a 5.8\% word error rate, a substantial improvement. 

Below, the teams behind the top 4 competition entries describe their solutions and the lessons learned. 

\section{Linderman Lab: 4th Place Entry}

We submitted two models to the competition. We initially focused on adapting deep state space models (SSM) for the purpose of neural speech decoding \cite{smith2022simplified,gu2023mamba}. The best performing deep SSM matched the CTC loss and phoneme error rate of the baseline GRU but it achieved a 11.05\% word error rate, which did not improve upon the baseline. We then tested whether modifications we made to the training procedure for optimizing the deep SSMs would improve training of the baseline RNN. This resulted in a baseline RNN model with a 9.22\% word error rate that we submitted to the competition, which improved on the original baseline RNN performance. Finally, after the competition we explored additional architectural modifications to the baseline RNN. This resulted in a new model with an 8.00\% word error rate. In the following sections, we describe our deep SSM results and the modifications to the baseline RNN. 

\subsection{Deep state space model (Mamba)}
Mamba is a recently developed deep state space model that has drawn much attention from the machine learning community due to its effectiveness and efficiency in modeling long-range sequence data \cite{gu2023mamba}.
To explore the application of Mamba on the speech BCI task, we implemented a bidirectional variant of Mamba as a drop-in replacement for the gated recurrent units (GRUs) in the PyTorch RNN baseline code \cite{schiff2024caduceus}.

We performed a hyperparameter search over all hyperparameters including the number of Mamba layers, the dimension of the Mamba state sizes, and the learning rate scheduler. While the best Mamba model matched the RNN performance in terms of validation CTC loss and phoneme error rate, the final word error rate was 1.5\% behind the RNN baseline.

We next speculate on reasons why the deep SSM models did not meet or improve upon the baseline RNN performance. First, it could be that the speech decoding task does not significantly depend on long-range information. Furthermore, the RNN receives inputs that are stacked across a window of history; this stacking provides direct paths between inputs at previous time points and the current state and may further decrease the need for an architecture designed to learn long-range sequential dependencies. Next, it could be that further modifications to the Mamba architecture and training scheme may be necessary to filter out fluctuations in the neural features that are not helpful for decoding intended speech. 

\subsection{Modified training of baseline RNN}
\label{subsec:modified_rnn}

We experimented with three different techniques to improve the training process of the baseline RNN.
First, we employed a step-wise learning rate decay, which scaled down the learning rate of the RNN training process by a factor of 10 after 7,500 iterations (out of 10,000 total training iterations).
This is based on our observation that the loss of the baseline RNN plateaus at around 7,500 iterations of training.
Next, we applied a regularization to the connectionist temporal classification loss to encourage emitting phoneme labels over blank labels \cite{yu2021fastemit}.
We employed this regularization based on our observation that the trained models favor blank labels over phoneme labels.
Lastly, we used coordinated dropout \cite{keshtkaran2019enabling} (or speckled masking), a dropout strategy that masks out random parts of the input neural data at every batch to help prevent the model from overfitting.
Through applications to the neural latent benchmark \cite{pei2021neural}, we found this strategy to be effective in modeling neural data with sequence models. Training the baseline RNN with these three techniques together resulted in our RNN entry that achieved a 9.22\% word error rate. 

\begin{figure}
  \centering
  \includegraphics[width=0.5\textwidth]{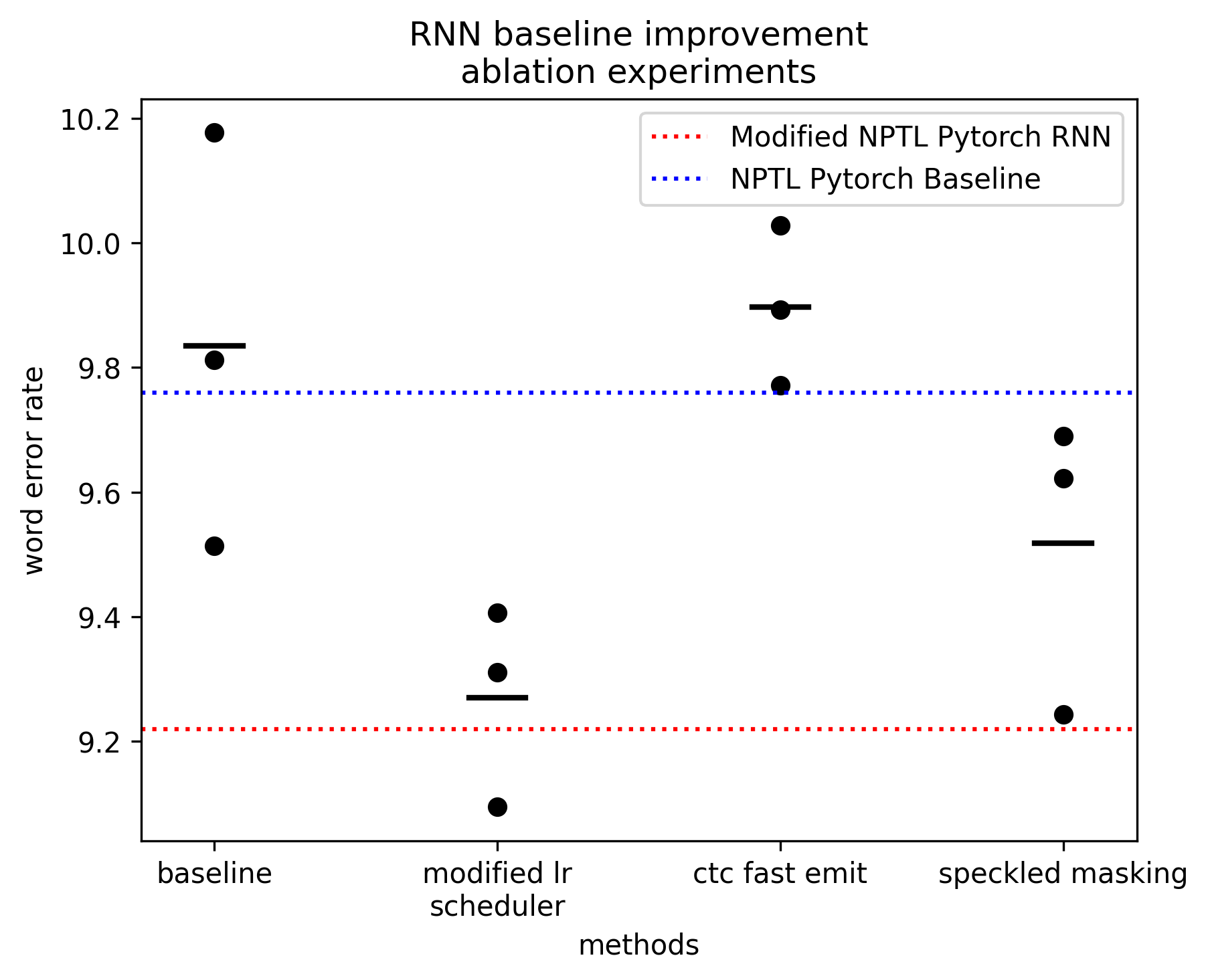}
  \caption{\textbf{Linderman Lab Baseline RNN Ablation Experiments}}
  \label{fig:linderman_lab_rnn_ablation}
\end{figure}

We then performed ablation experiments to see which of these techniques contributed to the improvement in the baseline RNN performance.
The results of the ablation experiments are shown in Figure \ref{fig:linderman_lab_rnn_ablation}.
The leftmost column is the baseline RNN, the second is the baseline RNN with a step-wise learning rate decay, the third is the baseline RNN with the FastEmit regularization, and the last column is the baseline RNN with coordinated dropout.
For each column, we trained three separate models with different seeds. The WERs for each fit are shown as black dots in each column and the mean for each column is shown as a horizontal black line. We noted two additional baselines on this figure: the WER of the baseline RNN (blue dotted line) and our RNN competition entry (red dotted line). 
This ablation study shows that the modification in the learning rate scheduler is effective in improving the performance, where the best model reached a WER of 9.095\%.
Coordinated dropout (speckled masking) also appeared to improve the performance. While the FastEmit regularization did not improve performance, we note that the regularization was designed to encourage faster label emission and therefore decreased response latency. Exploring whether the phoneme label timing changed when this regularization was included is a potential future analysis.  

\subsection{Baseline RNN with additional normalization achieves improved performance}
After the competition, we continued exploring improvements to the baseline RNN’s performance.
In particular, we experimented with employing layer normalization, a method commonly used to stabilize and improve the training of deep neural networks \cite{ba2016layer}.
We incorporated a stack of layers composed of layer normalization, dropout, a linear transformation, and a nonlinear activation after the RNN.
Additionally, drawing from the insights from our ablation studies in Section \ref{subsec:modified_rnn}, we employed a linear learning rate decay, which gradually reduced the learning rate from 0.025 to 0.0005 over 20,000 training iterations and speckled masking probability of 0.3.
Together, these modifications to the training of the baseline RNN resulted in a significant improvement, achieving an 8.00\% word error rate.

\section{LISA: 3rd Place Entry}
Our Large Language Model Integrated Scoring Adjustment (LISA) focused on two primary areas to improve the word error rate (WER) of brain-to-text decoding: ensembling and fine-tuned LLM rescoring \cite{benster2024cross}. By leveraging predictions from an ensemble of models and passing them through a fine-tuned language model, we achieved significant improvements in transcription accuracy.

We used multiple training runs of the PyTorch RNN baseline code with different random seeds to create diverse output predictions. By averaging or combining these outputs, we aimed to capture more accurate sentence-level transcriptions, which individual models might miss. This diversity improved the performance substantially. Each seed had similiar word error rate (WER) of 14.6\% on average (e.g. the first five baseline models trained had 15.1\%, 14.6\%, 14.4\%, 14.2\%, 14.6\%, and 14.4\% after beam search on the test partition). LISA without finetuning improved this to 13.7\% for 10 models x top1 beams. Following finetuning, LISA reached 8.9\% on the competition partition, and was the first submission to beat the PyTorch RNN baseline performance.

We found that performance was highly sensitive to the particular prompt. Undesired preambles and hallucinations were the largest drivers of WER, occasionally increasing over the baseline. We address this by prompting the model to choose the transcript that is most accurate. Empirically, this improved compliance to the task, without reducing the models capabilities to synthesize new sentences that were not a candidate. Finetuning further eliminated these sources of errors.

\section{TeamCyber: 2nd Place Entry}

\subsection{Optimizing the RNN training process}
We began by testing whether modifying the architecture of the baseline RNN neural decoder will improve performance. Specifically, we tested one-dimensional convolutional neural network (CNN), including temporal convolution network \cite{bai2018tcn}, either as a pre-processing step before the RNN or as a replacement for the RNN. Although some models incorporating the one-dimensional CNN as a pre-processing step achieved a low phoneme error rate (PER) of approximately 15\% on the validation set, these models exhibited higher word error rates (WER) compared to the baseline RNN model when combined with standard language model decoding (using 5-gram and OPT6.7B). This suggested to us that (i) baseline RNN architecture is quite powerful, and (ii) training neural decoders by minimizing the PER might not necessarily lead to models with low WER, observations that other participants have also noted.

Given the challenges associated with modifying the architecture of the neural decoder, we next turned to improving the RNN training process. Specifically, we experimented with different optimizers and learning rate schedulers. The original model employed Adam with a linear learning rate decay. In contrast, we trained two models using stochastic gradient descent (SGD) with momentum—one with Nesterov momentum and the other without. For the learning rate decay, we used a step-wise decay, similar to the 4th place entry, where we started the learning rate from 0.1 and decreased it by one tenth every 4,000 steps for model 1 or after 5,000 steps for model 2.

It is interesting to note that Adam builds upon SGD with momentum by incorporating adaptive learning rate for each parameter, making it a more general optimizer. Thus, we wondered why SGD with momentum, a simpler optimizer, did well in our case. A previous comprehensive experiment \cite{choi2019optimizers} suggested that if one has the computational budget to tune all the hyperparameters of an optimizer, performance typically aligns with this inclusion relationship SGD $\subseteq$ SGD with momentum $\subseteq$ Adam. Thus, if we tune all the hyperparameters of Adam (learning rate, $\beta_1$, $\beta_2$, and $\varepsilon$) thoroughly , Adam will likely perform similarly or even better. However, in real-world cases where training resources are limited and not all hyperparameters could be exhaustively tested, a simpler optimizer with less hyperparameters (such as SGD with momentum we used) may still be effective.

\subsection{A simple model ensembling}
\begin{figure}[!htbp]
  \centering
  \includegraphics[width=0.85\textwidth]{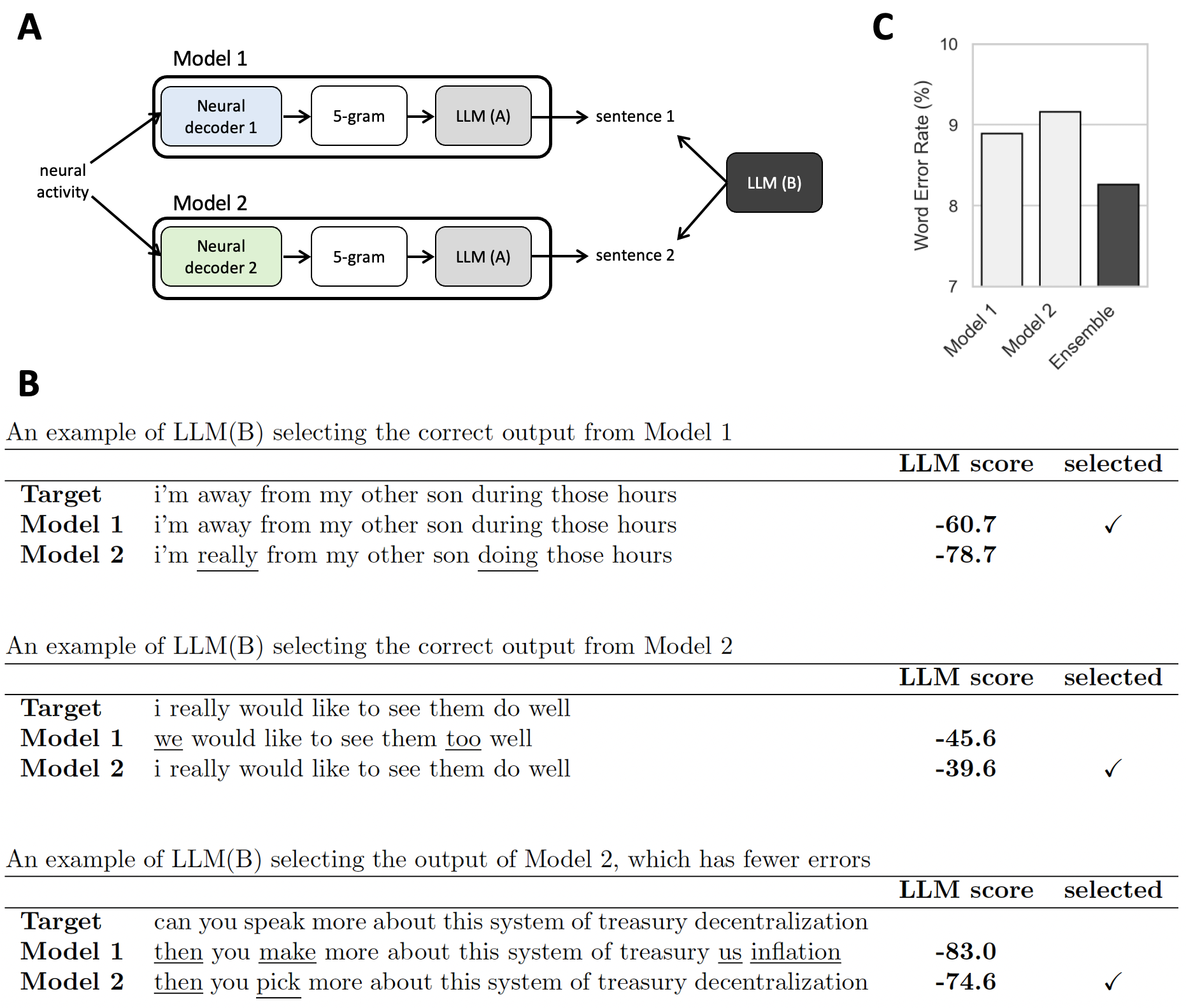}
  \caption{A simple model ensembling used by TeamCyber. \textbf{A}: Schematic of the model ensembling process. Two separate neural decoders combined with language model decoding (5-gram and LLM(A)) produced two distinct sentences. A separate LLM, referred to as LLM(B), was used to select the sentence with the higher score. We used OPT6.7B for LLM(A) and Llama2-7B-chat for LLM(B). \textbf{B}: Example output sentences from the two models and the resulting ensemble output. Word errors are underlined. \textbf{C}: Word error rate (WER) on the competition's held-out data.}
  \label{fig:TeamCyber_ensembling}
\end{figure}

For the language model decoding part, we included a model ensembling to combine the outputs of multiple models to improve the prediction, similar to the approach taken by other participants\cite{benster2024cross}. We trained two neural decoders with identical architectures but slightly different training parameters, as mentioned in the previous section. The outputs of these decoders were passed through the regular 5-gram and LLM decoding step (using OPT6.7B) leading to two predicted sentences for a given neural activity (Figure \ref{fig:TeamCyber_ensembling}A). We then used another LLM that evaluated the scores of these two sentences, and selected the sentence with the higher score. Preliminary tests indicated that LLMs optimized for dialogues are particularly effective for this purpose, leading us to select Llama2-7b-chat \cite{touvron2023llama2}.

Examination of the outputs of this model ensembling confirmed that this step was indeed helpful (Figure \ref{fig:TeamCyber_ensembling}B). If only one of the two models predicted a correct sentence, the second LLM often chose the correct one. If neither of sentences were correct, the second LLM often picked the sentence that had fewer word errors.  While the WER on the held-out dataset was 8.89\% and 9.16\% for the two individual models, the final output after model ensembling achieved a WER of 8.26\%\ (Figure \ref{fig:TeamCyber_ensembling}C). Our PyTorch Lightning training code and model checkpoints are available on this \href{https://github.com/CIBR-Okubo-Lab/speechBCI_2024}{GitHub repository}.

\section{DConD-LIFT: 1st Place Entry}
\subsection{Neural decoding with a context-aware prior}
Neural activity recorded from the motor cortex represents the oral facial movements that generate articulation. Decoding these neural signals into well-organized articulation units (phonemes) provides a foundation for translating neural signals into words and sentences. However, neural signals controlling continuous muscle movements have much higher granularity compared to the 40 discrete phoneme classes. It is highly likely that beyond encoding single phonemes, neural signals also encode the transitions between phonemes. For example, neural signals that produce movements from `AA' to `D' can differ from signals generating transitions from `IY' to `D'.
Based on this assumption, we propose \textbf{D}ivide-\textbf{Co}nquer-\textbf{N}eural-\textbf{D}ecoder (DCoND) to decode neural signals into context-dependent diphones instead of single phonemes \cite{li2024dcond}. The diphone classes include transitions between any two phonemes, resulting in approximately 1,600 classes—40 times more than single phoneme classes. To address this increased complexity, in DCoND we divide each phoneme recognition problem into 40 sub-problems, considering the transitions between the preceding phonemes and the current phoneme. Then the model is trained to predict the likelihood of each possible transition between phonemes. In the conquer step, we marginalize over the preceding phoneme to obtain the likelihood of the current phoneme.
Our results show that these steps reduce the Phoneme Error Rate from 16.62\% to 15.34\%. This improvement indicates that by incorporating a decoding strategy with stronger correlation between input (neural signals) and output (diphones), the neural decoder learns representations more consistent with neural coding strategies, resulting in better phoneme decoding accuracy.

\subsection{Phoneme-to-text conversion with large language models}
We applied large language models to estimate intended sentences from the decoded phonemes with three steps:

\begin{enumerate}
    \item Generation of potential phoneme-to-sentence transcription hypotheses using an n-gram model (5-gram)
    \item Re-scoring the transcriptions with OPT (Large Language Model)
    \item Merging an ensemble of re-scored transcriptions generated by ten RNN decoders independently, each using steps 1-2.
\end{enumerate}

For the first two steps, we followed the baseline method using 5-gram + OPT \cite{willett2023high}. The third step merges 10 sentence candidates output from an ensemble of 10 independent RNN decoders, where steps 1 and 2 have been applied to each RNN separately. These ten candidates are then merged into a single hypothesis using either a fine-tuned GPT-3.5 model, or GPT-3.5 with in-context learning.

While the final merging step is similar to \cite{benster2024cross}, it is unique in that it inputs both the text transcriptions and the decoded phonemes into GPT-3.5 (as opposed to the candidate text alone). The phoneme input allows GPT-3.5 to access the original neural decoder outputs, enabling potential correction of any errors introduced by the n-gram sentence generation and OPT transcription re-scoring procedures.
The Word Error Rate (WER) obtained through steps 1 and 2 alone with a single DCoND RNN was approximately 8.09\%. After implementing ensembling and merging using GPT-3.5 (DCoND-LIFT), the WER was reduced to 5.77\%. While in-context learning (DCoND-LI) proved less effective than fine-tuning, it still improved WER to 7.29\%. Additional ablation studies examining the contribution of the divide-conquer decoding strategy and different LLM components are detailed in \cite{li2024dcond}.

\section{Discussion}
The improvement that generated the largest gain in performance was the idea of using an ensemble of neural decoders to generate a diverse set of sentence hypotheses, and then using a fine-tuned large language model to merge these hypotheses into a finalized sentence. This idea was first introduced by Benster et al. \cite{benster2024cross} and quickly inspired other entries.  

One surprise was the difficulty entrants encountered when trying to improve upon the RNN architecture from the baseline solution. Although transformers \cite{vaswani2017attention} have long since surpassed recurrent neural networks for many machine learning tasks, they did not seem to improve performance here. One issue could simply be that more optimization is required, given that the RNN architecture and training approach has now been optimized for years to perform well at decoding neural activity \cite{willett2023high, willett2021high, sussillo2016making, deo2024brain, metzger2023high}. However, another possibility could be that RNNs are particularly well-suited for the neural decoding problem. For this task, it is likely that neural information related to each phoneme is local in time to that phoneme, and does not extend widely across the time series. In this case, RNNs, which naturally implement this prior, may be well-suited. Deep state space models \cite{smith2022simplified} and transformers, which are designed to model long-range relationships in sequential data, may not be as efficient at learning this task given small amounts of data. Transformers in particular have historically been applied to very large datasets in order to reach their optimal performance. In the future, when larger neural datasets are available, revisiting the comparison between the models could be advantageous.

One limitation of the baseline approach is its two-stage nature, consisting of a separate neural decoder and language modeling step. On the one hand, this may be an efficient approach for small datasets. On the other hand, inability to optimize the algorithm end-to-end could hurt performance, and also can lead to dissociations in the performance of the RNN and the performance of the whole system. For example, we have routinely observed that lower phoneme error rates from the RNN do not always lead to lower word error rates when the phoneme logits are decoded with the language model. Future work may aim to merge the two stages in a single neural network model (e.g., \cite{feng2024towards}).

We are optimistic about the outlook of continuing to improve performance further via algorithm improvements. Undoubtedly, performance will improve as more data is collected. The benchmark dataset consists of 10,000 sentences; in the near term, we believe at least 100,000 sentences should be feasible. Second, large language models will continue to improve, and larger datasets may also facilitate end-to-end models that combine language modeling with neural decoding. Third, tweaks to modern machine learning architectures (transformers, deep state space models) may allow better transfer to neural data and improve upon the RNN baseline. 

\section{Acknowledgments}
DCoND team (JL,TL,MC,ES) acknowledge the support of HDR Institute: Accelerated AI Algorithms for Data-Driven Discovery (A3D3) National Science Foundation grant PHY-2117997. and the departments of Applied Mathematics (ES) and Electrical and Computer Engineering (ES, JL, TL, MC) at the University of Washington.

\bibliographystyle{unsrt}  
\bibliography{references}  

\begin{thebibliography}{10}

\bibitem{card2024accurate}
Nicholas~S Card, Maitreyee Wairagkar, Carrina Iacobacci, Xianda Hou, Tyler Singer-Clark, Francis~R Willett, Erin~M Kunz, Chaofei Fan, Maryam Vahdati~Nia, Darrel~R Deo, et~al.
\newblock An accurate and rapidly calibrating speech neuroprosthesis.
\newblock {\em New England Journal of Medicine}, 391(7):609--618, 2024.

\bibitem{metzger2023high}
Sean~L Metzger, Kaylo~T Littlejohn, Alexander~B Silva, David~A Moses, Margaret~P Seaton, Ran Wang, Maximilian~E Dougherty, Jessie~R Liu, Peter Wu, Michael~A Berger, et~al.
\newblock A high-performance neuroprosthesis for speech decoding and avatar control.
\newblock {\em Nature}, 620(7976):1037--1046, 2023.

\bibitem{willett2023high}
Francis~R Willett, Erin~M Kunz, Chaofei Fan, Donald~T Avansino, Guy~H Wilson, Eun~Young Choi, Foram Kamdar, Matthew~F Glasser, Leigh~R Hochberg, Shaul Druckmann, et~al.
\newblock A high-performance speech neuroprosthesis.
\newblock {\em Nature}, 620(7976):1031--1036, 2023.

\bibitem{graves2006connectionist}
Alex Graves, Santiago Fern{\'a}ndez, Faustino Gomez, and J{\"u}rgen Schmidhuber.
\newblock Connectionist temporal classification: labelling unsegmented sequence data with recurrent neural networks.
\newblock In {\em Proceedings of the 23rd international conference on Machine learning}, pages 369--376, 2006.

\bibitem{smith2022simplified}
Jimmy~TH Smith, Andrew Warrington, and Scott~W Linderman.
\newblock Simplified state space layers for sequence modeling.
\newblock {\em arXiv preprint arXiv:2208.04933}, 2022.

\bibitem{gu2023mamba}
Albert Gu and Tri Dao.
\newblock Mamba: Linear-time sequence modeling with selective state spaces.
\newblock {\em arXiv preprint arXiv:2312.00752}, 2023.

\bibitem{schiff2024caduceus}
Yair Schiff, Chia-Hsiang Kao, Aaron Gokaslan, Tri Dao, Albert Gu, and Volodymyr Kuleshov.
\newblock Caduceus: Bi-directional equivariant long-range dna sequence modeling.
\newblock {\em arXiv preprint arXiv:2403.03234}, 2024.

\bibitem{yu2021fastemit}
Jiahui Yu, Chung-Cheng Chiu, Bo~Li, Shuo-yiin Chang, Tara~N Sainath, Yanzhang He, Arun Narayanan, Wei Han, Anmol Gulati, Yonghui Wu, et~al.
\newblock Fastemit: Low-latency streaming asr with sequence-level emission regularization.
\newblock In {\em ICASSP 2021-2021 IEEE International Conference on Acoustics, Speech and Signal Processing (ICASSP)}, pages 6004--6008. IEEE, 2021.

\bibitem{keshtkaran2019enabling}
Mohammad~Reza Keshtkaran and Chethan Pandarinath.
\newblock Enabling hyperparameter optimization in sequential autoencoders for spiking neural data.
\newblock {\em Advances in neural information processing systems}, 32, 2019.

\bibitem{pei2021neural}
Felix Pei, Joel Ye, David Zoltowski, Anqi Wu, Raeed~H Chowdhury, Hansem Sohn, Joseph~E O'Doherty, Krishna~V Shenoy, Matthew~T Kaufman, Mark Churchland, et~al.
\newblock Neural latents benchmark'21: evaluating latent variable models of neural population activity.
\newblock {\em arXiv preprint arXiv:2109.04463}, 2021.

\bibitem{ba2016layer}
Jimmy~Lei Ba.
\newblock Layer normalization.
\newblock {\em arXiv preprint arXiv:1607.06450}, 2016.

\bibitem{benster2024cross}
Tyler Benster, Guy Wilson, Reshef Elisha, Francis~R Willett, and Shaul Druckmann.
\newblock A cross-modal approach to silent speech with llm-enhanced recognition.
\newblock {\em arXiv preprint arXiv:2403.05583}, 2024.

\bibitem{bai2018tcn}
Shaojie Bai, J.~Zico Kolter, and Vladlen Koltun.
\newblock An empirical evaluation of generic convolutional and recurrent networks for sequence modeling.
\newblock {\em arXiv preprint arXiv:1803.01271}, 2018.

\bibitem{choi2019optimizers}
Dami Choi, Christopher~J. Shallue, Zachary Nado, Jaehoon Lee, Chris~J. Maddison, and George~E. Dahl.
\newblock On empirical comparisons of optimizers for deep learning.
\newblock {\em arXiv preprint arXiv:1910.05446}, 2019.

\bibitem{touvron2023llama2}
Hugo Touvron, Louis Martin, Kevin Stone, Peter Albert, Amjad Almahairi, Yasmine Babaei, Nikolay Bashlykov, Soumya Batra, Prajjwal Bhargava, Shruti Bhosale, Dan Bikel, Lukas Blecher, Cristian~Canton Ferrer, Moya Chen, Guillem Cucurull, David Esiobu, Jude Fernandes, Jeremy Fu, Wenyin Fu, Brian Fuller, Cynthia Gao, Vedanuj Goswami, Naman Goyal, Anthony Hartshorn, Saghar Hosseini, Rui Hou, Hakan Inan, Marcin Kardas, Viktor Kerkez, Madian Khabsa, Isabel Kloumann, Artem Korenev, Punit~Singh Koura, Marie-Anne Lachaux, Thibaut Lavril, Jenya Lee, Diana Liskovich, Yinghai Lu, Yuning Mao, Xavier Martinet, Todor Mihaylov, Pushkar Mishra, Igor Molybog, Yixin Nie, Andrew Poulton, Jeremy Reizenstein, Rashi Rungta, Kalyan Saladi, Alan Schelten, Ruan Silva, Eric~Michael Smith, Ranjan Subramanian, Xiaoqing~Ellen Tan, Binh Tang, Ross Taylor, Adina Williams, Jian~Xiang Kuan, Puxin Xu, Zheng Yan, Iliyan Zarov, Yuchen Zhang, Angela Fan, Melanie Kambadur, Sharan Narang, Aurelien Rodriguez, Robert Stojnic, Sergey Edunov, and Thomas
  Scialom.
\newblock Llama 2: Open foundation and fine-tuned chat models.
\newblock {\em arXiv preprint arXiv:2307.09288}, 2023.

\bibitem{li2024dcond}
Jingyuan Li, Trung Le, Chaofei Fan, Mingfei Chen, and Eli Shlizerman.
\newblock Brain-to-text decoding with context-aware neural representations and large language models.
\newblock {\em arXiv preprint arXiv:2411.10657}, 2024.

\bibitem{vaswani2017attention}
Ashish Vaswani, Noam Shazeer, Niki Parmar, Jakob Uszkoreit, Llion Jones, Aidan~N Gomez, {\L}ukasz Kaiser, and Illia Polosukhin.
\newblock Attention is all you need.
\newblock {\em Advances in neural information processing systems}, 30, 2017.

\bibitem{willett2021high}
Francis~R Willett, Donald~T Avansino, Leigh~R Hochberg, Jaimie~M Henderson, and Krishna~V Shenoy.
\newblock High-performance brain-to-text communication via handwriting.
\newblock {\em Nature}, 593(7858):249--254, 2021.

\bibitem{sussillo2016making}
David Sussillo, Sergey~D Stavisky, Jonathan~C Kao, Stephen~I Ryu, and Krishna~V Shenoy.
\newblock Making brain--machine interfaces robust to future neural variability.
\newblock {\em Nature communications}, 7(1):13749, 2016.

\bibitem{deo2024brain}
Darrel~R Deo, Francis~R Willett, Donald~T Avansino, Leigh~R Hochberg, Jaimie~M Henderson, and Krishna~V Shenoy.
\newblock Brain control of bimanual movement enabled by recurrent neural networks.
\newblock {\em Scientific Reports}, 14(1):1598, 2024.

\bibitem{feng2024towards}
Sheng Feng, Heyang Liu, Yu~Wang, and Yanfeng Wang.
\newblock Towards an end-to-end framework for invasive brain signal decoding with large language models.
\newblock {\em arXiv preprint arXiv:2406.11568}, 2024.

\end{thebibliography}

\end{document}